\documentclass[twoside,11pt]{article}

\usepackage{jmlr2e}
\usepackage{graphicx}
\usepackage{float}
\usepackage{listings}
\usepackage{mathtools}
\usepackage{pgfplots}
\usepackage[super]{nth}

\pgfkeys{/pgf/declare function={arcsinh(\x) = ln(\x + sqrt(\x^2+1));}}
\pgfplotsset{compat=1.14}

\ShortHeadings{m-arcsinh: An Efficient and Reliable Function for SVM and MLP in scikit-learn}{Luca Parisi, \emph{PhD, MBA Candidate}}
\firstpageno{1}

\begin{document}

\title{m-arcsinh: An Efficient and Reliable Function for SVM and MLP in scikit-learn}

\author{\name Luca Parisi \email luca.parisi@ieee.org \\
       \addr Coventry, United Kingdom \\
       PhD in Machine Learning for Clinical Decision Support Systems \\ 
       MBA Candidate with Artificial Intelligence Specialism
       }

\editor{}

\maketitle


\begin{abstract}
This paper describes the 'm-arcsinh', a modified ('m-') version of the inverse hyperbolic sine function ('arcsinh'). Kernel and activation functions enable Machine Learning (ML)-based algorithms, such as Support Vector Machine (SVM) and Multi-Layer Perceptron (MLP), to learn from data in a supervised manner.
m-arcsinh, implemented in the open source Python library 'scikit-learn', is hereby presented as an efficient and reliable kernel and activation function for SVM and MLP respectively. 
Improvements in reliability and speed to convergence in classification tasks on fifteen (N = 15) datasets available from scikit-learn and the University California Irvine (UCI) Machine Learning repository are discussed. 
Experimental results demonstrate the overall competitive classification performance of both SVM and MLP, achieved via the proposed function. 
This function is compared to gold standard kernel and activation functions, demonstrating its overall competitive reliability regardless of the complexity of the classification tasks involved.
\end{abstract}

\vskip 0.1in

\begin{keywords}
  Kernel, Activation, Support Vector Machine, Multi-Layer Perceptron, Scikit-learn
\end{keywords}


\vskip 0.3in

\section{Introduction}

\vskip 0.15in

Despite theoretical advances in both kernel and activation functions respectively for optimal separating hyperplane (OSH)-based classifiers, such as the Support Vector Machine (SVM)~\citep{cortes1995support}, and Artificial Neural Networks (ANN), e.g., the Multi-Layer Perceptron (MLP)~\citep{rumelhart1986learning}, usable, reproducible and replicable functions for both the SVM and the MLP have remained limited and confined to two limited sets of functions deemed as 'gold standard'. Both sets of functions have been made freely accessible in the open source Python library named 'scikit-learn'~\citep{scikit-learn} for Machine Learning, under the related 'MLPClassifier' and the 'SVC' (Support Vector Classifier) classes, which respectively implement the MLP and SVM for classification. The availability of these functions to the public has made it possible for ecosystems of organisations across academia and industry to leverage these assets for various purposes and applications, ranging from teaching aids to practical user-centred implementations~\citep{buitinck2013api}.

Nevertheless, both sets of functions are not only limited with respect to their number, but also they may not always lead to reliable outcomes when applied for classification, e.g., causing slow or lack of convergence~\citep{vert2006consistency} ~\citep{jacot2018neural}, due to trapping at local minima~\citep{parisi2020novel}. Moreover, both sets of functions are mutually exclusive~\citep{scikit-learn}, given the challenge in deriving a mathematical function that can be used as a kernel function for SVM and as an activation function for MLP concurrently. The only similarity between them is the presence of the 'sigmoid' kernel function in SVM and its modified version named 'tanh' or 'hyperbolic tangent sigmoid'~\citep{lin2003study}, which has an extended range ([-1, +1], as opposed to [0, +1]) and a stronger gradient due to steeper derivatives, making it more suitable for ANN, such as the MLP, rather than OSH-based classifiers, e.g., SVM. 

Therefore, there is an increasing need for additional open source kernel and activation functions, which reach convergence faster, avoiding trapping at \emph{local minima}, are more stable and can also be used across multiple algorithms. Entirely written in Python and made freely available in 'scikit-learn'~\citep{scikit-learn} for both the 'MLPClassifier' and the 'SVC' classes, the proposed hyperbolic function is demonstrated as a competitive function with respect to gold standard functions, which suits both kernel and activation functions' requirements, thus being computationally efficient and reliable.

Thanks to its liberal license, it has been widely distributed as a part of the free software Python library 'scikit-learn'~\citep{scikit-learn}, and it is available for use for both academic research and commercial purposes.\\


\vskip 0.3in

\section{Methods}

\vskip 0.15in

\subsection{Datasets used from scikit-learn and the UCI ML repository}

\vskip 0.15in

The following datasets from \emph{scikit-learn} were used in the experiments described and discussed in this study:

\begin{itemize}

\item 'Breast cancer Wisconsin (diagnostic)' dataset~\citep{Wolberg:1995}, having 30 characteristics of cell \emph{nuclei} from 569 digitised images of a fine needle aspirate of breast masses, to detect whether they correspond to either malignant or benign breast cancer;
\item 'LFW people' dataset~\citep{LFWTech}~\citep{LFWTechUpdate}, which has 13,233 JPEG photos of 5,749 famous people collected from the Internet, each of which is composed of 5,828 features, to identify the individual appearing on each photo;
\item 'Iris' dataset~\citep{anderson1936species}~\citep{fisher1936use}, which has three species of one-hundred and fifty (N=150) 'Iris' flowers to be classified based on four features describing their petals and sepals;
\item 'Handwritten Digits' dataset~\citep{alpaydin1998opt}, to recognise handwritten digits (from 0 to 9), given about 180 images per class (1,797 images in total) and 64 features per each image;
\item 'Wine' dataset~\citep{Forina:1991}, which has 13 features derived from 178 measurements obtained via a chemical analysis of wines grown in the same region in Italy by three different cultivators, to understand whether such measurements and different constituents correspond to one of three types of wine (59 measurements for the first type, 71 for the second type, 48 for the third type);
\item 'Olivetti faces' dataset~\citep{Roweis:2017} with (only) 10 different 64x64 images of the faces of 40 different subjects - to be identified via classification - taken between April 1992 and April 1994 at the 'AT and T' Laboratories Cambridge. Such photos were taken against a dark homogeneous background at different times, with various lighting, facial expressions (open/closed eyes, smiling/not smiling) and details (glasses/no glasses). Subjects were in an upright, frontal position, with little side movement at time;
 
\end{itemize}

Moreover, the following datasets from \emph{The University California Irvine (UCI) ML repository} were used for additional evaluation in this study:

\begin{itemize}

\item 'Optical Recognition of Handwritten Digits' (‘OptDigits’) datasets~\citep{kaynak1995methods}, to recognise handwritten digits (from 0 to 9), given 5,620 images in total and 64 features per each image, from 43 people, 30 of which contributed to the training data partition and the remaining 13 to the partition for testing:

\begin{itemize}
\item \href{https://archive.ics.uci.edu/ml/machine-learning-databases/optdigits/optdigits.tra}{training data partition} ('optdigits.tra' file);
\item \href{https://archive.ics.uci.edu/ml/machine-learning-databases/optdigits/optdigits.tes}{testing data partition} ('optdigits.tes' file);
\end{itemize}

\item \href{https://archive.ics.uci.edu/ml/machine-learning-databases/00519/heart_failure_clinical_records_dataset.csv}{‘Heart failure clinical records’ dataset}~\citep{chicco2020machine}, to predict whether a patient was deceased during the follow-up period, based on 13 clinical features from medical records of 299 patients who had heart failure;

\item \href{https://archive.ics.uci.edu/ml/machine-learning-databases/parkinsons/parkinsons.data}{‘Parkinson’s’ dataset}~\citep{little2007exploiting}, which has 23 features corresponding to 195 biomedical voice measurements from 31 people, 23 with Parkinson's disease (PD), to help in detecting PD from speech signals; 

\item \href{https://archive.ics.uci.edu/ml/machine-learning-databases/haberman/haberman.data}{‘Haberman survival’ dataset}~\citep{Lim:1999}, with three features (age of patient at time of operation; patient's year of operation; number of positive axillary nodes detected) to predict whether 306 patients who had undergone surgery for breast cancer would have died within 5 years of follow up or survived for longer;

\item ‘SPECTF’ dataset~\citep{Cios:2001}, which has 267 images collected via a cardiac Single Proton Emission Computed Tomography (SPECT), describing whether each patient has a physiological or pathophysiological heart based on 44 features:

\begin{itemize}
\item \href{https://archive.ics.uci.edu/ml/machine-learning-databases/spect/SPECTF.train}{training data partition} ('SPECTF.train' file), with 80 images;
\item \href{https://archive.ics.uci.edu/ml/machine-learning-databases/spect/SPECTF.test}{testing data partition} ('SPECTF.test' file), which has 187 images;
\end{itemize}

\item \href{http://archive.ics.uci.edu/ml/machine-learning-databases/statlog/german/german.data-numeric}{‘German Statlog credit data’}~\citep{Hofmann:1994}, to identify whether a customer is associated with a good or bad credit risk based on 20 features;

\item ‘Pen-based handwritten digits recognition’ dataset~\citep{alpaydin1998pen}, to recognise handwritten digits (from 0 to 9), drawn on a WACOM PL-100V pressure sensitive tablet with an integrated LCD display and a cordless stylus, based on 250 images from 44 writers:

\begin{itemize}
\item \href{https://archive.ics.uci.edu/ml/machine-learning-databases/pendigits/pendigits.tra}{training data partition} ('pendigits.tra' file), with images from 30 writers;
\item \href{https://archive.ics.uci.edu/ml/machine-learning-databases/pendigits/pendigits.tes}{testing data partition} ('pendigits.tes' file), which has images from the remaining 14 writers;
\end{itemize}

\item \href{https://archive.ics.uci.edu/ml/machine-learning-databases/00422/wifi_localization.txt}{‘Wireless Indoor Localization’ dataset}~\citep{bhatt2005fuzzy}~\citep{rohra2017user}, which has 7 features characterising the strength of a Wi-Fi signal observed on a smartphone in indoor spaces to identify if an individual was in one of four rooms;

\item \href{https://archive.ics.uci.edu/ml/machine-learning-databases/00451/dataR2.csv}{'Breast Cancer Coimbra’ dataset}~\citep{patricio2018using}, with 10 clinical features, including anthropometric data and parameters collected via haematological analysis, measured for 64 patients with breast cancer and 52 healthy controls to identify the presence or absence of breast cancer.

\end{itemize}

\vskip 0.15in

\subsection{Baseline SVM and MLP models and hyperparameters}

\vskip 0.15in

As the purpose of this study is not to devise the most optimised, best-performing classifier for any of the classification tasks involved in 2.1, but, instead, to develop a novel computationally efficient and reliable kernel and activation function and evaluate it against the two sets of gold standard functions available in the Python library 'scikit-learn'~\citep{scikit-learn} under the 'MLPClassifier' and the 'SVC' classes, baseline SVM and MLP models were used with the following hyperparameters for all classification tasks in 2.1:

\begin{itemize}

\item MLP-related hyperparameters:
\begin{itemize}
\item \emph{'random\_state'} = 1;
\item \emph{'max\_iter'} =300, where 'max\_iter' is the maximum number of iterations.
\end{itemize}

Listing 1 provides the snippet of code in Python to use an MLP with different activation functions available in 'scikit-learn'~\citep{scikit-learn}, including the novel 'm-arcsinh'.

\lstset{language=Python}
\lstset{frame=lines}
\lstset{caption={MLP with different activation functions available in 'scikit-learn'~\citep{scikit-learn}, including the proposed 'm-arcsinh'.}}
\lstset{label={lst:code_direct_1}}
\lstset{basicstyle=\footnotesize}
\begin{lstlisting}

from sklearn.neural_network import MLPClassifier

for activation in ('identity', 'logistic', 'tanh', 'relu', 'm-arcsinh'):
    classifier =  MLPClassifier(activation=activation, 
    random_state=1, max_iter=300)

\end{lstlisting}

\item SVM-related hyperparameters:
\begin{itemize}
\item \emph{'gamma'} = 0.001, where 'gamma' is the kernel coefficient for the ‘rbf’, ‘poly’ and ‘sigmoid’ kernel functions;
\item \emph{'random\_state'} = 13;
\item \emph{'class\_weight'} ='balanced', setting the parameter C by adjusting the weights to be inversely proportional to the class frequencies in the input data.

\end{itemize}

Listing 2 provides the snippet of code in Python to use an SVM with different kernel functions available in 'scikit-learn'~\citep{scikit-learn}, including the novel 'm-arcsinh'.

\lstset{language=Python}
\lstset{frame=lines}
\lstset{caption={SVM with different kernel functions available in 'scikit-learn'~\citep{scikit-learn}, including the proposed 'm-arcsinh'.}}
\lstset{label={lst:code_direct_2}}
\lstset{basicstyle=\footnotesize}
\begin{lstlisting}

from sklearn import svm

for kernel in ('linear', 'poly', 'rbf', 'sigmoid', 'm_arcsinh'):
    classifier = svm.SVC(kernel=kernel, gamma=0.001, random_state=13,
    class_weight='balanced')

\end{lstlisting}

\end{itemize}

Where the data were not already provided in two separate partitions for training and testing (see 2.1), the datasets were split via \emph{'train\_test\_split'} in 'scikit-learn'~\citep{scikit-learn} from \emph{'sklearn.model\_selection'} as follows, without randomisation (\emph{'shuffle'=False}): 

\begin{itemize}

\item 70\% of the data was selected for training, whilst the remaining 30\% for testing for the 'Handwritten Digits' dataset;
\item 80\% for training, 20\% for testing for the 'Statlog', 'Olivetti faces', 'Parkinson's', 'Wi-Fi localization', 'Breast Cancer Coimbra', 'Haberman' and 'Heart Failure' datasets;
\item 75\% for training, 25\% for testing for the 'LFW people' dataset.

\end{itemize}

\vskip 0.15in

\subsection{m-arcsinh: A new kernel and activation function}

For a function to be both generalised as a kernel and activation function for SVM and MLP, it has to be able to 1) maximise the margin width in SVM and 2) improve discrimination of input data into target classes via a transfer mechanism of appropriately extended range for MLP. Two functions that satisfy the two above-mentioned requirements are the linear kernel for SVM and tanh for MLP. Nevertheless, whilst the linear kernel is not suitable in MLP to leverage gradient descent training appropriately in presence of non-linearly separable data, the tanh function has an extended range with sigmoidal behaviour for SVM to maximise the margin width reliably with such data.

Thus, a novel function was devised to be suitable for both SVM and MLP concurrently by leveraging a weighted interaction effect between the hyperbolic nature of the inverse hyperbolic sine function ('arcsinh'), suitable for MLP, and the slightly non-linear characteristic of the squared root function, appropriate for SVM. With higher weight (1/3) given to the 'arcsinh' and a slightly lower one (1/4) to the square root function, hence satisfying both the above-mentioned requirements 1) and 2) concurrently, the following modified (m-) arcsinh (m-arcsinh) was derived:

\vspace{1em}
\begin{math}
arcsinh(x) \times \frac{1}{3} \times \frac{1}{4} \times \sqrt{\left|x\right|} = arcsinh(x) \times \frac{1}{12} \times  \sqrt{\left|x\right|}\hspace{13.5em}(1)
\end{math}
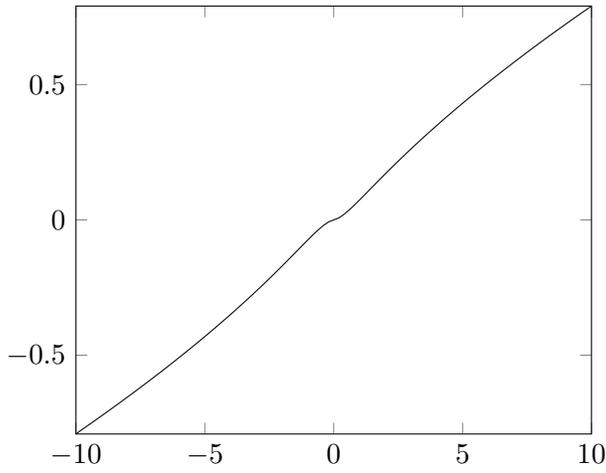
\vspace{2em}

\begin{tikzpicture}
\begin{axis}[enlargelimits=false]
\addplot [domain=-10:10, samples=101,unbounded coords=jump]{(arcsinh(x)/3)*(sqrt(abs(x))/4)};
\end{axis}
\end{tikzpicture}

\vspace{2em}

The derivative of m-arcsinh can be expressed as:

\vspace{1em}
\begin{math}
{\sqrt{\left|x\right|}} \times \frac{1}{12\times\sqrt{x^2+1}}  +\  \frac{x\ \times arcsinh(x)}{24\times\left|x\right|^\frac{3}{2}\ }
\end{math}
\hspace{36.5em}(2)
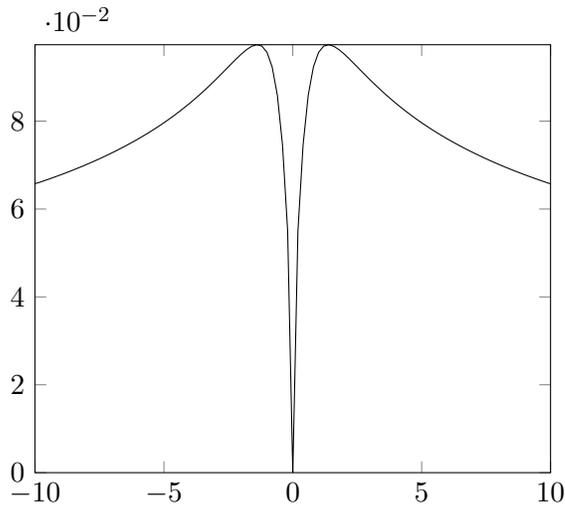
\vspace{2em}

\begin{tikzpicture}
\begin{axis}[enlargelimits=false]
\addplot [domain=-10:10, samples=101,unbounded coords=jump]{sqrt(abs(x))/(12*sqrt(x^2+1)) + (x*arcsinh(x))/(24*abs(x)^(3/2)))};
\end{axis}
\end{tikzpicture}

\vspace{2em}

Listing 3 provides the snippet of code in Python that implements the proposed m-arcsinh function as a kernel for an SVM classifier or 'SVC' in 'scikit-learn'~\citep{scikit-learn}.

\lstset{language=Python}
\lstset{frame=lines}
\lstset{caption={Using the m-arcsinh function as a kernel for an SVM classifier or 'SVC' in 'scikit-learn'~\citep{scikit-learn}.}}
\lstset{label={lst:code_direct_3}}
\lstset{basicstyle=\footnotesize}
\begin{lstlisting}

import numpy as np
from sklearn import svm

# X is the numpy ndarray of the inputs to classify, 
# Y is the numpy ndarray of the target classes.

def m_arcsinh(X, Y):

    return np.dot((1/3*np.arcsinh(X))*(1/4*np.sqrt(np.abs(X))),
    (1/3*np.arcsinh(Y.T))*(1/4*np.sqrt(np.abs(Y.T))))

classifier = svm.SVC(kernel=m_arcsinh, gamma=0.001, random_state=13,
class_weight='balanced')

\end{lstlisting}

Listing 4 provides the snippet of code in Python that implements the proposed m-arcsinh function (1) as an activation and its derivative (2) for an MLP classifier or 'MLPClassifier' in 'scikit-learn'~\citep{scikit-learn}.

\lstset{language=Python}
\lstset{frame=lines}
\lstset{caption={Using the m-arcsinh function as a kernel for an MLP classifier or 'MLPClassifier' in 'scikit-learn'~\citep{scikit-learn}.}}
\lstset{label={lst:code_direct_4}}
\lstset{basicstyle=\footnotesize}
\begin{lstlisting}

import numpy as np
from sklearn.neural_network import MLPClassifier


def m_arcsinh(X):
    """Compute the m-arcsinh hyperbolic function in place.

    Parameters
    ----------
    X: {array-like, sparse matrix}, shape (n_samples, n_features)
        The input data.

    Returns
    -------
    X_new: {array-like, sparse matrix}, shape (n_samples, n_features)
        The transformed data.
    """
    	
    return (1/3*np.arcsinh(X))*(1/4*np.sqrt(np.abs(X)))
    
    
def inplace_m_arcsinh_derivative(Z, delta):
    """Apply the derivative of the hyperbolic m-arcsinh function.

    It exploits the fact that the derivative is a simple function 
    of the output value from the hyperbolic m-arcsinh.

    Parameters
    ----------
    Z : {array-like, sparse matrix}, shape (n_samples, n_features)
        The data which were output from the hyperbolic 
        m-arcsinh activation function during the forward pass.

    delta : {array-like}, shape (n_samples, n_features)
         The back-propagated error signal to be modified in place.
    """
    delta *= (np.sqrt(np.abs(Z))/(12*np.sqrt(Z**2+1)) 
    + (Z*np.arcsinh(Z))/(24*np.abs(Z)**(3/2)))


classifier =  MLPClassifier(activation='m_arcsinh',
random_state=1, max_iter=300)

\end{lstlisting}

\vskip 0.15in

\subsection{Performance evaluation}

\vskip 0.15in

The accuracy of the SVM and MLP using different kernel and activation functions respectively, as described in 2.3 and 2.4 on the datasets outlined in 2.1, was evaluated via the \emph{'accuracy\_score'} available in 'scikit-learn'~\citep{scikit-learn} from \emph{'sklearn.metrics'}. 
The reliability of such classifiers was assessed via the weighted average of the precision, recall and F1-score computed via the \emph{'classification\_report'}, also available in 'scikit-learn'~\citep{scikit-learn} from \emph{'sklearn.metrics'}.
\\
To understand what classification accuracy and reliability are, and how they can be evaluated, please refer to the following studies:~\citep{parisi2018decision},~\citep{parisi2018feature},~\citep{parisi2020novel},~\citep{parisi2020evolutionary}.

Moreover, the computational cost of the classifiers, to quantify the impact of using different kernel and activation functions, was assessed via the training time in seconds. Experiments were run on an AMD E2-9000 Radeon R2 processor, 1.8 GHz and 4 GB DDR4 RAM.


\section{Results}

\vskip 0.15in

Experimental results demonstrate the competitiveness of the proposed m-arcsinh kernel and activation function for SVM and MLP respectively, as being accurate, reliable, and computationally efficient, with the following classification performance and training time:
\begin{itemize}
\item For the MLP: 
\begin{itemize}
\item The best classification performance on 10 out of 15 datasets evaluated (Tables 2, 3, 5-7, and Tables 7, 9, 10, 12-14 in the Appendix).
\item The \nth{2} highest classification performance on 4 out of 15 datasets evaluated (Tables 1 and 4, and Tables 8 and 11 in the Appendix).
\item The fastest training time on 2 out of 15 datasets assessed (Tables 13 and 15 in the Appendix).
\item The best classification performance and the fastest training time on 2 out of 15 datasets assessed (Tables 13 and 15 in the Appendix).
\end{itemize}

\item For the SVM: 
\begin{itemize}
\item The best classification performance on 2 out of 15 datasets assessed (Table 2, Table 14 in the Appendix).
\item The \nth{2} highest classification performance on 5 out of 15 datasets evaluated (Tables 1,3,4, and Tables 12 and 13 in the Appendix).
\item The fastest training time on 7 out of 15 datasets assessed (Tables 1, 3, 6, 7, and Tables 9, 11, 12 in the Appendix).
\item The \nth{2} highest classification performance and the fastest training time on 2 out of 15 datasets evaluated (Tables 1 and 3).
\end{itemize}
\end{itemize}

\vskip 0.145in

\newpage


\textbf{Table 1.} Results on performance evaluation of baseline (non-optimised) Support Vector Machine (SVM) and Multi-Layer Perceptron (MLP) in scikit-learn with different kernel and activation functions respectively, including the proposed m-arcsinh function. The performance of such classifiers was evaluated on the ‘Breast cancer Wisconsin (diagnostic)’ dataset~\citep{Wolberg:1995} available in scikit-learn. 

\begin{table}[H]
\resizebox{\textwidth}{!}{%
\begin{tabular}{lllllll}
\textbf{Classifier} &
  \textbf{Kernel   function} &
  \textbf{Training   time (s)} &
  \textbf{Accuracy   (0-1)} &
  \textbf{\begin{tabular}[c]{@{}l@{}}Weighted   precision  \\    \\ (0-1)\end{tabular}} &
  \textbf{\begin{tabular}[c]{@{}l@{}}Weighted   recall \\    \\ (0-1)\end{tabular}} &
  \textbf{\begin{tabular}[c]{@{}l@{}}Weighted   F1-score \\    \\ (0-1)\end{tabular}} \\
SVM & m-arcsinh   (this study) & 0.007   & 0.97 & 0.97 & 0.97 & 0.97 \\
SVM & RBF                      & 0.017   & 0.92 & 0.92 & 0.92 & 0.92 \\
SVM & Linear                   & 1.312   & 0.98 & 0.98 & 0.98 & 0.98 \\
SVM & Poly                     & 311.706 & 0.98 & 0.98 & 0.98 & 0.98 \\
SVM & Sigmoid                  & 0.012   & 0.39 & 0.15 & 0.39 & 0.21 \\
MLP & m-arcsinh   (this study) & 9.830   & 0.91 & 0.92 & 0.91 & 0.91 \\
MLP & Identity                 & 3.124   & 0.92 & 0.92 & 0.92 & 0.92 \\
MLP & Logistic                 & 3.638   & 0.92 & 0.92 & 0.92 & 0.92 \\
MLP & tanh                     & 3.568   & 0.90 & 0.90 & 0.90 & 0.90 \\
MLP & ReLU                     & 3.132   & 0.92 & 0.92 & 0.92 & 0.92
\end{tabular}%
}
\end{table}


\textbf{Table 2.} Results on performance evaluation of baseline (non-optimised) Support Vector Machine (SVM) and Multi-Layer Perceptron (MLP) in scikit-learn with different kernel and activation functions respectively, including the proposed m-arcsinh function. The performance of such classifiers was evaluated on the ‘OptDigits’ dataset~\citep{kaynak1995methods} available at the University California Irvine (UCI) Machine Learning repository. 

\begin{table}[H]
\resizebox{\textwidth}{!}{%
\begin{tabular}{lllllll}
\textbf{Classifier} &
  \textbf{Function} &
  \textbf{Training   time (s)} &
  \textbf{Accuracy   (0-1)} &
  \textbf{\begin{tabular}[c]{@{}l@{}}Weighted   precision  \\    \\ (0-1)\end{tabular}} &
  \textbf{\begin{tabular}[c]{@{}l@{}}Weighted   recall \\    \\ (0-1)\end{tabular}} &
  \textbf{\begin{tabular}[c]{@{}l@{}}Weighted   F1-score \\    \\ (0-1)\end{tabular}} \\
SVM & m-arcsinh   (this study) & 0.232 & 0.97 & 0.97 & 0.97 & 0.97 \\
SVM & RBF                      & 0.525 & 0.98 & 0.98 & 0.98 & 0.98 \\
SVM & Linear                   & 0.175 & 0.96 & 0.96 & 0.96 & 0.96 \\
SVM & Poly                     & 0.180 & 0.97 & 0.98 & 0.97 & 0.97 \\
SVM & Sigmoid                  & 2.384 & 0.71 & 0.75 & 0.71 & 0.72 \\
MLP & m-arcsinh   (this study) & 53.586 & 0.98 & 0.98 & 0.98 & 0.98 \\
MLP & Identity                 & 9.572 & 0.98 & 0.98 & 0.98 & 0.98 \\
MLP & Logistic                 & 27.457 & 0.98 & 0.98 & 0.98 & 0.98 \\
MLP & tanh                     & 15.750 & 0.98 & 0.98 & 0.98 & 0.98 \\
MLP & ReLU                     & 14.254 & 0.98 & 0.98 & 0.98 & 0.98
\end{tabular}%
}
\end{table}


\textbf{Table 3.} Results on performance evaluation of baseline (non-optimised) Support Vector Machine (SVM) and Multi-Layer Perceptron (MLP) in scikit-learn with different kernel and activation functions respectively, including the proposed m-arcsinh function. The performance of such classifiers was evaluated on the ‘LFW people’ dataset~\citep{LFWTech}~\citep{LFWTechUpdate} available in scikit-learn.

\begin{table}[H]
\resizebox{\textwidth}{!}{%
\begin{tabular}{lllllll}
\textbf{Classifier} &
  \textbf{Function} &
  \textbf{Training   time (s)} &
  \textbf{Accuracy   (0-1)} &
  \textbf{\begin{tabular}[c]{@{}l@{}}Weighted   precision  \\    \\ (0-1)\end{tabular}} &
  \textbf{\begin{tabular}[c]{@{}l@{}}Weighted   recall \\    \\ (0-1)\end{tabular}} &
  \textbf{\begin{tabular}[c]{@{}l@{}}Weighted   F1-score \\    \\ (0-1)\end{tabular}} \\
SVM & m-arcsinh   (this study) & 0.083 & 0.83 & 0.84 & 0.83 & 0.83 \\
SVM & RBF                      & 0.508 & 0.85 & 0.87 & 0.85 & 0.85 \\
SVM & Linear                   & 0.230 & 0.78 & 0.80 & 0.78 & 0.79 \\
SVM & Poly                     & 0.483 & 0.05 & 0.00 & 0.05 & 0.00 \\
SVM & Sigmoid                  & 0.570 & 0.82 & 0.83 & 0.82 & 0.82 \\
MLP & m-arcsinh   (this study) & 7.101 & 0.86 & 0.86 & 0.86 & 0.86 \\
MLP & Identity                 & 6.225 & 0.84 & 0.84 & 0.84 & 0.83 \\
MLP & Logistic                 & 7.892 & 0.85 & 0.85 & 0.85 & 0.84 \\
MLP & tanh                     & 5.562 & 0.84 & 0.84 & 0.84 & 0.84 \\
MLP & ReLU                     & 4.755 & 0.84 & 0.84 & 0.84 & 0.83
\end{tabular}%
}
\end{table}

\newpage


\textbf{Table 4.} Results on performance evaluation of baseline (non-optimised) Support Vector Machine (SVM) and Multi-Layer Perceptron (MLP) in scikit-learn with different kernel and activation functions respectively, including the proposed m-arcsinh function. The performance of such classifiers was evaluated on the ‘Iris’ dataset~\citep{anderson1936species}~\citep{fisher1936use} available in scikit-learn. 

\begin{table}[H]
\resizebox{\textwidth}{!}{%
\begin{tabular}{lllllll}
\textbf{Classifier} &
  \textbf{Function} &
  \textbf{Training   time (s)} &
  \textbf{Accuracy   (0-1)} &
  \textbf{\begin{tabular}[c]{@{}l@{}}Weighted   precision  \\    \\ (0-1)\end{tabular}} &
  \textbf{\begin{tabular}[c]{@{}l@{}}Weighted   recall \\    \\ (0-1)\end{tabular}} &
  \textbf{\begin{tabular}[c]{@{}l@{}}Weighted   F1-score \\    \\ (0-1)\end{tabular}} \\
SVM & m-arcsinh   (this study) & 0.002 & 0.93 & 0.95 & 0.93 & 0.93 \\
SVM & RBF                      & 0.003 & 0.63 & 0.43 & 0.63 & 0.50 \\
SVM & Linear                   & 0.001 & 0.97 & 0.97 & 0.97 & 0.97 \\
SVM & Poly                     & 0.003 & 0.33 & 0.11 & 0.33 & 0.17 \\
SVM & Sigmoid                  & 0.002 & 0.50 & 0.43 & 0.50 & 0.40 \\
MLP & m-arcsinh   (this study) & 2.357 & 0.90 & 0.93 & 0.90 & 0.90 \\
MLP & Identity                 & 0.707 & 0.93 & 0.95 & 0.93 & 0.93 \\
MLP & Logistic                 & 1.553 & 0.93 & 0.95 & 0.93 & 0.93 \\
MLP & tanh                     & 0.939 & 0.93 & 0.95 & 0.93 & 0.93 \\
MLP & ReLU                     & 1.344 & 0.93 & 0.95 & 0.93 & 0.93
\end{tabular}%
}
\end{table}


\textbf{Table 5.} Results on performance evaluation of baseline (non-optimised) Support Vector Machine (SVM) and Multi-Layer Perceptron (MLP) in scikit-learn with different kernel and activation functions respectively, including the proposed m-arcsinh function. The performance of such classifiers was evaluated on the ‘Heart failure clinical records’ dataset~\citep{chicco2020machine} available at the University California Irvine (UCI) Machine Learning repository.

\begin{table}[H]
\resizebox{\textwidth}{!}{%
\begin{tabular}{lllllll}
\textbf{Classifier} &
  \textbf{Function} &
  \textbf{Training   time (s)} &
  \textbf{Accuracy   (0-1)} &
  \textbf{\begin{tabular}[c]{@{}l@{}}Weighted   precision  \\    \\ (0-1)\end{tabular}} &
  \textbf{\begin{tabular}[c]{@{}l@{}}Weighted   recall \\    \\ (0-1)\end{tabular}} &
  \textbf{\begin{tabular}[c]{@{}l@{}}Weighted   F1-score \\    \\ (0-1)\end{tabular}} \\
SVM & m-arcsinh   (this study) & 1.285              & 0.88 & 0.90 & 0.88 & 0.89 \\
SVM & RBF                      & 0.007              & 0.78 & 0.61 & 0.78 & 0.69 \\
SVM & Linear                   & 48.287             & 0.85 & 0.89 & 0.85 & 0.86 \\
SVM & Poly                     & Did   not converge & N/A  & N/A  & N/A  & N/A  \\
SVM & Sigmoid                  & 0.005              & 0.78 & 0.61 & 0.78 & 0.69 \\
MLP & m-arcsinh   (this study) & 0.013              & 0.78 & 0.61 & 0.78 & 0.69 \\
MLP & Identity                 & 0.023              & 0.78 & 0.61 & 0.78 & 0.69 \\
MLP & Logistic                 & 0.011              & 0.78 & 0.61 & 0.78 & 0.69 \\
MLP & tanh                     & 0.010              & 0.78 & 0.61 & 0.78 & 0.69 \\
MLP & ReLU                     & 0.016              & 0.78 & 0.61 & 0.78 & 0.69
\end{tabular}%
}
\end{table}


\textbf{Table 6.} Results on performance evaluation of baseline (non-optimised) Support Vector Machine (SVM) and Multi-Layer Perceptron (MLP) in scikit-learn with different kernel and activation functions respectively, including the proposed m-arcsinh function. The performance of such classifiers was evaluated on the ‘Parkinson’s’ dataset available at the University California Irvine (UCI) Machine Learning repository~\citep{little2007exploiting}.

\begin{table}[H]
\resizebox{\textwidth}{!}{%
\begin{tabular}{lllllll}
\textbf{Classifier} &
  \textbf{Function} &
  \textbf{Training   time (s)} &
  \textbf{Accuracy   (0-1)} &
  \textbf{\begin{tabular}[c]{@{}l@{}}Weighted   precision  \\    \\ (0-1)\end{tabular}} &
  \textbf{\begin{tabular}[c]{@{}l@{}}Weighted   recall \\    \\ (0-1)\end{tabular}} &
  \textbf{\begin{tabular}[c]{@{}l@{}}Weighted   F1-score \\    \\ (0-1)\end{tabular}} \\
SVM & m-arcsinh   (this study) & 0.005 & 0.79 & 0.78 & 0.79 & 0.79 \\
SVM & RBF                      & 0.005 & 0.77 & 0.82 & 0.77 & 0.78 \\
SVM & Linear                   & 0.182 & 0.87 & 0.92 & 0.87 & 0.88 \\
SVM & Poly                     & 5.911 & 0.82 & 0.83 & 0.82 & 0.82 \\
SVM & Sigmoid                  & 0.005 & 0.77 & 0.59 & 0.77 & 0.67 \\
MLP & m-arcsinh   (this study) & 0.008 & 0.77 & 0.59 & 0.77 & 0.67 \\
MLP & Identity                 & 0.009 & 0.77 & 0.59 & 0.77 & 0.67 \\
MLP & Logistic                 & 0.016 & 0.77 & 0.59 & 0.77 & 0.67 \\
MLP & tanh                     & 0.016 & 0.77 & 0.59 & 0.77 & 0.67 \\
MLP & ReLU                     & 0.001 & 0.77 & 0.59 & 0.77 & 0.67
\end{tabular}%
}
\end{table}

\newpage


\textbf{Table 7.} Results on performance evaluation of baseline (non-optimised) Support Vector Machine (SVM) and Multi-Layer Perceptron (MLP) in scikit-learn with different kernel and activation functions respectively, including the proposed m-arcsinh function. The performance of such classifiers was evaluated on the ‘Haberman survival’ dataset~\citep{Lim:1999} available at the University California Irvine (UCI) Machine Learning repository.

\begin{table}[H]
\resizebox{\textwidth}{!}{%
\begin{tabular}{lllllll}
\textbf{Classifier} &
  \textbf{Function} &
  \textbf{Training   time (s)} &
  \textbf{Accuracy   (0-1)} &
  \textbf{\begin{tabular}[c]{@{}l@{}}Weighted   precision  \\    \\ (0-1)\end{tabular}} &
  \textbf{\begin{tabular}[c]{@{}l@{}}Weighted   recall \\    \\ (0-1)\end{tabular}} &
  \textbf{\begin{tabular}[c]{@{}l@{}}Weighted   F1-score \\    \\ (0-1)\end{tabular}} \\
SVM & m-arcsinh   (this study) & 0.004 & 0.77 & 0.77 & 0.77 & 0.77 \\
SVM & RBF                      & 0.004 & 0.79 & 0.78 & 0.79 & 0.79 \\
SVM & Linear                   & 0.014 & 0.79 & 0.77 & 0.79 & 0.78 \\
SVM & Poly                     & 0.010 & 0.77 & 0.77 & 0.77 & 0.77 \\
SVM & Sigmoid                  & 0.005 & 0.82 & 0.68 & 0.82 & 0.74 \\
MLP & m-arcsinh   (this study) & 1.219 & 0.76 & 0.77 & 0.76 & 0.76 \\
MLP & Identity                 & 1.095 & 0.76 & 0.77 & 0.76 & 0.76 \\
MLP & Logistic                 & 0.957 & 0.76 & 0.77 & 0.76 & 0.76 \\
MLP & tanh                     & 0.912 & 0.76 & 0.77 & 0.76 & 0.76 \\
MLP & ReLU                     & 0.974 & 0.76 & 0.77 & 0.76 & 0.76
\end{tabular}%
}
\end{table}


\vskip 0.3in

\section{Discussion}

\vskip 0.15in

As demonstrated by the competitive results obtained on the 15 datasets evaluated, especially those in Tables 2, 3, 5-7 and Tables 9, 10, 12-14 in the Appendix for the MLP, Tables 1, 3-5, and Tables 12-14 in the Appendix for the SVM, the m-arcsinh is deemed a suitable kernel and activation function for SVM and MLP respectively.
\\
In fact, its reliability was high, as quantified via appropriate metrics in 2.4, and better than some gold standard functions, e.g., considering Table 1 with the F1-score of the SVM using m-arcsinh being 0.97 as opposed to that of the SVM using RBF or sigmoid being 0.92 and 0.21 respectively. Moreover, its computational efficiency was generally high, e.g., considering Table 1 with the training time of the SVM leveraging m-arcsinh being only 0.007 seconds as compared to that of the SVM using linear or polynomial kernel being 1.312 and 311.706 seconds.
\\
Therefore, the m-arcsinh demonstrates that it is possible for a function to be generalised as a kernel and activation function concurrently and the mathematical formulation of such a function does not have to be sophisticated at all.
As a reliable and computationally efficient function, the m-arcsinh is thus deemed a new gold standard kernel and activation function for SVM and MLP, freely available in scikit-learn. 


\section{Conclusion}

m-arcsinh in scikit-learn provides a function in supervised ML that serves as a kernel for SVM and activation for MLP for classification. It is a fast and stable kernel and activation function, thus being a competitive candidate amongst the available gold standard functions for SVM and MLP in scikit-learn. 
Since it is made freely available, open source, on the Python and scikit-learn ecosystems, it adds to the choices that both academia and industry can have when selecting or optimising for kernel and activation functions for SVM and MLP respectively. 
Importantly, the proposed algorithm, being computationally efficient and reliable, and written in a high-level programming language (Python), can be leveraged as a part of ML-based pipelines in specific use cases, wherein high accuracy and reliability need to be achieved, whilst powerful computational hardware may not always be available, such as in the healthcare sector, including small clinics.
Future work involves adapting this function to benefit deep neural networks too.


\acks{This research did not receive any specific grant from funding agencies in the public, commercial, or not-for-profit sectors.}


\newpage

\appendix
\section*{Appendix}



In this appendix, as mentioned in the 'Results' section of this article, further results are provided in support of the proposed m-arcsinh kernel and activation function for SVM and MLP, implemented in Python and made freely available in scikit-learn, in Tables 8-15 on datasets from both scikit-learn and the UCI ML repository.
\\

\textbf{Table 8.} Results on performance evaluation of baseline (non-optimised) Support Vector Machine (SVM) and Multi-Layer Perceptron (MLP) in scikit-learn with different kernel and activation functions respectively, including the proposed m-arcsinh function. The performance of such classifiers was evaluated on the ‘Handwritten Digits’ dataset~\citep{alpaydin1998opt} available in scikit-learn.

\begin{table}[H]
\resizebox{\textwidth}{!}{%
\begin{tabular}{lllllll}
\textbf{Classifier} &
  \textbf{Function} &
  \textbf{Training   time (s)} &
  \textbf{Accuracy   (0-1)} &
  \textbf{\begin{tabular}[c]{@{}l@{}}Weighted   precision \\    \\ (0-1)\end{tabular}} &
  \textbf{\begin{tabular}[c]{@{}l@{}}Weighted   recall\\    \\ (0-1)\end{tabular}} &
  \textbf{\begin{tabular}[c]{@{}l@{}}Weighted   F1-score\\    \\ (0-1)\end{tabular}} \\
SVM & m-arcsinh   (this study) & 0.037  & 0.95 & 0.95 & 0.95 & 0.95 \\
SVM & RBF                      & 0.116  & 0.97 & 0.97 & 0.97 & 0.97 \\
SVM & Linear                   & 0.033  & 0.93 & 0.93 & 0.93 & 0.93 \\
SVM & Poly                     & 0.043  & 0.95 & 0.95 & 0.95 & 0.95 \\
SVM & Sigmoid                  & 0.332  & 0.68 & 0.69 & 0.68 & 0.66 \\
MLP & m-arcsinh   (this study) & 28.650 & 0.92 & 0.92 & 0.92 & 0.92 \\
MLP & Identity                 & 5.452  & 0.91 & 0.91 & 0.91 & 0.91 \\
MLP & Logistic                 & 14.182 & 0.94 & 0.94 & 0.94 & 0.93 \\
MLP & tanh                     & 7.258  & 0.93 & 0.93 & 0.93 & 0.93 \\
MLP & ReLU                     & 7.834  & 0.92 & 0.92 & 0.92 & 0.92
\end{tabular}%
}
\end{table}


\textbf{Table 9.} Results on performance evaluation of baseline (non-optimised) Support Vector Machine (SVM) and Multi-Layer Perceptron (MLP) in scikit-learn with different kernel and activation functions respectively, including the proposed m-arcsinh function. The performance of such classifiers was evaluated on the ‘Wine’ dataset~\citep{Forina:1991} available in scikit-learn. 

\begin{table}[H]
\resizebox{\textwidth}{!}{%
\begin{tabular}{lllllll}
\textbf{Classifier} &
  \textbf{Function} &
  \textbf{Training   time (s)} &
  \textbf{Accuracy   (0-1)} &
  \textbf{\begin{tabular}[c]{@{}l@{}}Weighted   precision  \\    \\ (0-1)\end{tabular}} &
  \textbf{\begin{tabular}[c]{@{}l@{}}Weighted   recall \\    \\ (0-1)\end{tabular}} &
  \textbf{\begin{tabular}[c]{@{}l@{}}Weighted   F1-score \\    \\ (0-1)\end{tabular}} \\
SVM & m-arcsinh   (this study) & 0.003 & 0.89 & 0.89 & 0.89 & 0.89 \\
SVM & RBF                      & 0.003 & 0.72 & 0.77 & 0.72 & 0.72 \\
SVM & Linear                   & 0.162 & 0.94 & 0.95 & 0.94 & 0.94 \\
SVM & Poly                     & 0.094 & 0.97 & 0.97 & 0.97 & 0.97 \\
SVM & Sigmoid                  & 0.003 & 0.31 & 0.09 & 0.31 & 0.14 \\
MLP & m-arcsinh   (this study) & 0.008 & 0.72 & 0.77 & 0.72 & 0.72 \\
MLP & Identity                 & 0.016 & 0.72 & 0.77 & 0.72 & 0.72 \\
MLP & Logistic                 & 0.008 & 0.72 & 0.77 & 0.72 & 0.72 \\
MLP & tanh                     & 0.001 & 0.72 & 0.77 & 0.72 & 0.72 \\
MLP & ReLU                     & 0.008 & 0.72 & 0.77 & 0.72 & 0.72
\end{tabular}%
}
\end{table}

\newpage


\textbf{Table 10.} Results on performance evaluation of baseline (non-optimised) Support Vector Machine (SVM) and Multi-Layer Perceptron (MLP) in scikit-learn with different kernel and activation functions respectively, including the proposed m-arcsinh function. The performance of such classifiers was evaluated on the ‘SPECTF’ dataset~\citep{Cios:2001} available at the University California Irvine (UCI) Machine Learning repository.

\begin{table}[H]
\resizebox{\textwidth}{!}{%
\begin{tabular}{lllllll}
\textbf{Classifier} &
  \textbf{Function} &
  \textbf{Training   time (s)} &
  \textbf{Accuracy   (0-1)} &
  \textbf{\begin{tabular}[c]{@{}l@{}}Weighted   precision  \\    \\ (0-1)\end{tabular}} &
  \textbf{\begin{tabular}[c]{@{}l@{}}Weighted   recall \\    \\ (0-1)\end{tabular}} &
  \textbf{\begin{tabular}[c]{@{}l@{}}Weighted   F1-score \\    \\ (0-1)\end{tabular}} \\
SVM & m-arcsinh   (this study) & 0.004 & 0.91 & 0.91 & 0.91 & 0.91 \\
SVM & RBF                      & 0.003 & 0.98 & 0.98 & 0.97 & 0.97 \\
SVM & Linear                   & 0.004 & 1.00 & 1.00 & 1.00 & 1.00 \\
SVM & Poly                     & 0.003 & 1.00 & 1.00 & 1.00 & 1.00 \\
SVM & Sigmoid                  & 0.003 & 0.50 & 0.25 & 0.50 & 0.33 \\
MLP & m-arcsinh   (this study) & 0.047 & 0.54 & 0.76 & 0.54 & 0.41 \\
MLP & Identity                 & 0.080 & 0.54 & 0.76 & 0.54 & 0.41 \\
MLP & Logistic                 & 0.043 & 0.54 & 0.76 & 0.54 & 0.41 \\
MLP & tanh                     & 0.078 & 0.54 & 0.76 & 0.54 & 0.41 \\
MLP & ReLU                     & 0.096 & 0.54 & 0.76 & 0.54 & 0.41
\end{tabular}%
}
\end{table}


\textbf{Table 11.} Results on performance evaluation of baseline (non-optimised) Support Vector Machine (SVM) and Multi-Layer Perceptron (MLP) in scikit-learn with different kernel and activation functions respectively, including the proposed m-arcsinh function. The performance of such classifiers was evaluated on the ‘German Statlog credit data’~\citep{Hofmann:1994} available at the University California Irvine (UCI) Machine Learning repository.

\begin{table}[H]
\resizebox{\textwidth}{!}{%
\begin{tabular}{lllllll}
\textbf{Classifier} &
  \textbf{Function} &
  \textbf{Training   time (s)} &
  \textbf{Accuracy   (0-1)} &
  \textbf{\begin{tabular}[c]{@{}l@{}}Weighted   precision  \\    \\ (0-1)\end{tabular}} &
  \textbf{\begin{tabular}[c]{@{}l@{}}Weighted   recall \\    \\ (0-1)\end{tabular}} &
  \textbf{\begin{tabular}[c]{@{}l@{}}Weighted   F1-score \\    \\ (0-1)\end{tabular}} \\
SVM & m-arcsinh   (this study) & 0.023 & 0.70 & 0.74 & 0.69 & 0.71 \\
SVM & RBF                      & 0.043 & 0.71 & 0.75 & 0.71 & 0.73 \\
SVM & Linear                   & 0.931 & 0.71 & 0.75 & 0.71 & 0.73 \\
SVM & Poly                     & 0.093 & 0.72 & 0.75 & 0.72 & 0.73 \\
SVM & Sigmoid                  & 0.028 & 0.55 & 0.62 & 0.55 & 0.57 \\
MLP & m-arcsinh   (this study) & 12.895 & 0.79 & 0.78 & 0.79 & 0.78 \\
MLP & Identity                 & 1.328 & 0.79 & 0.78 & 0.79 & 0.78 \\
MLP & Logistic                 & 7.189 & 0.80 & 0.79 & 0.80 & 0.79 \\
MLP & tanh                     & 6.769 & 0.79 & 0.78 & 0.79 & 0.78 \\
MLP & ReLU                     & 5.459 & 0.79 & 0.77 & 0.79 & 0.78
\end{tabular}%
}
\end{table}


\textbf{Table 12.} Results on performance evaluation of baseline (non-optimised) Support Vector Machine (SVM) and Multi-Layer Perceptron (MLP) in scikit-learn with different kernel and activation functions respectively, including the proposed m-arcsinh function. The performance of such classifiers was evaluated on the ‘Olivetti faces’ dataset~\citep{Roweis:2017} available in scikit-learn.

\begin{table}[H]
\resizebox{\textwidth}{!}{%
\begin{tabular}{lllllll}
\textbf{Classifier} &
  \textbf{Function} &
  \textbf{Training   time (s)} &
  \textbf{Accuracy   (0-1)} &
  \textbf{\begin{tabular}[c]{@{}l@{}}Weighted   precision  \\    \\ (0-1)\end{tabular}} &
  \textbf{\begin{tabular}[c]{@{}l@{}}Weighted   recall \\    \\ (0-1)\end{tabular}} &
  \textbf{\begin{tabular}[c]{@{}l@{}}Weighted   F1-score \\    \\ (0-1)\end{tabular}} \\
SVM & m-arcsinh   (this study) & 0.143 & 0.91 & 0.91 & 0.91 & 0.90 \\
SVM & RBF                      & 1.452 & 0.31 & 0.28 & 0.31 & 0.27 \\
SVM & Linear                   & 1.124 & 0.99 & 0.99 & 0.99 & 0.99 \\
SVM & Poly                     & 1.071 & 0.85 & 0.85 & 0.85 & 0.83 \\
SVM & Sigmoid                  & 1.364 & 0.00 & 0.00 & 0.00 & 0.00 \\
MLP & m-arcsinh   (this study) & 105.341 & 0.75 & 0.78 & 0.75 & 0.75 \\
MLP & Identity                 & 109.109 & 0.75 & 0.78 & 0.75 & 0.75 \\
MLP & Logistic                 & 94.982 & 0.75 & 0.78 & 0.75 & 0.75 \\
MLP & tanh                     & 103.759 & 0.75 & 0.78 & 0.75 & 0.75 \\
MLP & ReLU                     & 104.581 & 0.75 & 0.78 & 0.75 & 0.75
\end{tabular}%
}
\end{table}

\newpage


\textbf{Table 13.} Results on performance evaluation of baseline (non-optimised) Support Vector Machine (SVM) and Multi-Layer Perceptron (MLP) in scikit-learn with different kernel and activation functions respectively, including the proposed m-arcsinh function. The performance of such classifiers was evaluated on the ‘Pen-based handwritten digits recognition’ dataset~\citep{alpaydin1998pen} available at the University California Irvine (UCI) Machine Learning repository.

\begin{table}[H]
\resizebox{\textwidth}{!}{%
\begin{tabular}{lllllll}
\textbf{Classifier} &
  \textbf{Function} &
  \textbf{Training   time (s)} &
  \textbf{Accuracy   (0-1)} &
  \textbf{\begin{tabular}[c]{@{}l@{}}Weighted   precision  \\    \\ (0-1)\end{tabular}} &
  \textbf{\begin{tabular}[c]{@{}l@{}}Weighted   recall \\    \\ (0-1)\end{tabular}} &
  \textbf{\begin{tabular}[c]{@{}l@{}}Weighted   F1-score \\    \\ (0-1)\end{tabular}} \\
SVM & m-arcsinh   (this study) & 0.728  & 0.99 & 0.99 & 0.99 & 0.99 \\
SVM & RBF                      & 2.922  & 1.00 & 1.00 & 1.00 & 1.00 \\
SVM & Linear                   & 4.651  & 0.99 & 0.99 & 0.99 & 0.99 \\
SVM & Poly                     & 0.196  & 1.00 & 1.00 & 1.00 & 1.00 \\
SVM & Sigmoid                  & 3.024  & 0.13 & 0.06 & 0.13 & 0.06 \\
MLP & m-arcsinh   (this study) & 17.736 & 1.00 & 1.00 & 1.00 & 1.00 \\
MLP & Identity                 & 18.692  & 1.00 & 1.00 & 1.00 & 1.00 \\
MLP & Logistic                 & 18.268 & 1.00 & 1.00 & 1.00 & 1.00 \\
MLP & tanh                     & 20.490 & 1.00 & 1.00 & 1.00 & 1.00 \\
MLP & ReLU                     & 19.362  & 1.00 & 1.00 & 1.00 & 1.00
\end{tabular}%
}
\end{table}


\textbf{Table 14.} Results on performance evaluation of baseline (non-optimised) Support Vector Machine (SVM) and Multi-Layer Perceptron (MLP) in scikit-learn with different kernel and activation functions respectively, including the proposed m-arcsinh function. The performance of such classifiers was evaluated on the ‘Wireless Indoor Localization’ dataset~\citep{bhatt2005fuzzy}~\citep{rohra2017user} available at the University California Irvine (UCI) Machine Learning repository.

\begin{table}[H]
\resizebox{\textwidth}{!}{%
\begin{tabular}{lllllll}
\textbf{Classifier} &
  \textbf{Function} &
  \textbf{Training   time (s)} &
  \textbf{Accuracy   (0-1)} &
  \textbf{\begin{tabular}[c]{@{}l@{}}Weighted   precision  \\    \\ (0-1)\end{tabular}} &
  \textbf{\begin{tabular}[c]{@{}l@{}}Weighted   recall \\    \\ (0-1)\end{tabular}} &
  \textbf{\begin{tabular}[c]{@{}l@{}}Weighted   F1-score \\    \\ (0-1)\end{tabular}} \\
SVM & m-arcsinh   (this study) & 0.022 & 0.99 & 0.99 & 0.99 & 0.99 \\
SVM & RBF                      & 0.015 & 0.99 & 0.99 & 0.99 & 0.99 \\
SVM & Linear                   & 0.021 & 0.99 & 0.99 & 0.99 & 0.99 \\
SVM & Poly                     & 0.038 & 0.99 & 0.99 & 0.99 & 0.99 \\
SVM & Sigmoid                  & 0.081 & 0.21 & 0.05 & 0.21 & 0.07 \\
MLP & m-arcsinh   (this study) & 5.870 & 0.98 & 0.98 & 0.98 & 0.98 \\
MLP & Identity                 & 4.767 & 0.98 & 0.98 & 0.98 & 0.98 \\
MLP & Logistic                 & 4.855 & 0.98 & 0.98 & 0.98 & 0.98 \\
MLP & tanh                     & 4.615 & 0.98 & 0.98 & 0.98 & 0.98 \\
MLP & ReLU                     & 6.343 & 0.98 & 0.98 & 0.98 & 0.98
\end{tabular}%
}
\end{table}


\textbf{Table 15.} Results on performance evaluation of baseline (non-optimised) Support Vector Machine (SVM) and Multi-Layer Perceptron (MLP) in scikit-learn with different kernel and activation functions respectively, including the proposed m-arcsinh function. The performance of such classifiers was evaluated on the ‘Breast Cancer Coimbra’ dataset~\citep{patricio2018using} available at the University California Irvine (UCI) Machine Learning repository.

\begin{table}[H]
\resizebox{\textwidth}{!}{%
\begin{tabular}{lllllll}
\textbf{Classifier} &
  \textbf{Function} &
  \textbf{Training   time (s)} &
  \textbf{Accuracy   (0-1)} &
  \textbf{\begin{tabular}[c]{@{}l@{}}Weighted   precision  \\    \\ (0-1)\end{tabular}} &
  \textbf{\begin{tabular}[c]{@{}l@{}}Weighted   recall \\    \\ (0-1)\end{tabular}} &
  \textbf{\begin{tabular}[c]{@{}l@{}}Weighted   F1-score \\    \\ (0-1)\end{tabular}} \\
SVM & m-arcsinh   (this study) & 0.005  & 0.71 & 0.71 & 0.71 & 0.71 \\
SVM & RBF                      & 0.003  & 0.71 & 0.72 & 0.71 & 0.70 \\
SVM & Linear                   & 0.726  & 0.75 & 0.75 & 0.75 & 0.75 \\
SVM & Poly                     & 83.401 & 0.79 & 0.79 & 0.79 & 0.79 \\
SVM & Sigmoid                  & 0.003  & 0.46 & 0.21 & 0.46 & 0.29 \\
MLP & m-arcsinh   (this study) & 0.001  & 0.46 & 0.21 & 0.46 & 0.29 \\
MLP & Identity                 & 0.017  & 0.46 & 0.21 & 0.46 & 0.29 \\
MLP & Logistic                 & 0.016  & 0.46 & 0.21 & 0.46 & 0.29 \\
MLP & tanh                     & 0.001  & 0.46 & 0.21 & 0.46 & 0.29 \\
MLP & ReLU                     & 0.001  & 0.46 & 0.21 & 0.46 & 0.29
\end{tabular}%
}
\end{table}

\newpage

\bibliography{ArXiv_paper_LP}

\begin{thebibliography}{29}
\providecommand{\natexlab}[1]{#1}
\providecommand{\url}[1]{\texttt{#1}}
\expandafter\ifx\csname urlstyle\endcsname\relax
  \providecommand{\doi}[1]{doi: #1}\else
  \providecommand{\doi}{doi: \begingroup \urlstyle{rm}\Url}\fi

\bibitem[Alpaydin and Alimoglu(1998)]{alpaydin1998pen}
E.~Alpaydin and F.~Alimoglu.
\newblock Pen-based recognition of handwritten digits data set - {UCI} machine
  learning repository, 1998.
\newblock URL
  \url{https://archive.ics.uci.edu/ml/datasets/Pen-Based+Recognition+of+Handwritten+Digits}.

\bibitem[Alpaydin and Kaynak(1998)]{alpaydin1998opt}
E.~Alpaydin and C.~Kaynak.
\newblock Optical recognition of handwritten digits data set - {UCI} machine
  learning repository, 1998.
\newblock URL
  \url{https://archive.ics.uci.edu/ml/datasets/Optical+Recognition+of+Handwritten+Digits}.

\bibitem[Anderson(1936)]{anderson1936species}
Edgar Anderson.
\newblock The species problem in iris.
\newblock \emph{Annals of the Missouri Botanical Garden}, 23\penalty0
  (3):\penalty0 457--509, 1936.

\bibitem[Bhatt(2005)]{bhatt2005fuzzy}
R~Bhatt.
\newblock Fuzzy-rough approaches for pattern classification: Hybrid measures,
  mathematical analysis, feature selection algorithms, decision tree
  algorithms, neural learning, and applications.
\newblock In \emph{Decision Tree Algorithms, Neural Learning, and
  Applications}. Amazon Books, 2005.

\bibitem[Buitinck et~al.(2013)Buitinck, Louppe, Blondel, Pedregosa, Mueller,
  Grisel, Niculae, Prettenhofer, Gramfort, Grobler, et~al.]{buitinck2013api}
Lars Buitinck, Gilles Louppe, Mathieu Blondel, Fabian Pedregosa, Andreas
  Mueller, Olivier Grisel, Vlad Niculae, Peter Prettenhofer, Alexandre
  Gramfort, Jaques Grobler, et~al.
\newblock Api design for machine learning software: experiences from the
  scikit-learn project.
\newblock \emph{arXiv preprint arXiv:1309.0238}, 2013.

\bibitem[Chicco and Jurman(2020)]{chicco2020machine}
Davide Chicco and Giuseppe Jurman.
\newblock Machine learning can predict survival of patients with heart failure
  from serum creatinine and ejection fraction alone.
\newblock \emph{BMC medical informatics and decision making}, 20\penalty0
  (1):\penalty0 16, 2020.

\bibitem[Cios et~al.(2001)Cios, Kurgan, and Goodenday]{Cios:2001}
K.~J. Cios, L.~A. Kurgan, and L.~S. Goodenday.
\newblock Spectf heart data set - {UCI} machine learning repository, 2001.
\newblock URL \url{https://archive.ics.uci.edu/ml/datasets/SPECTF+Heart}.

\bibitem[Cortes and Vapnik(1995)]{cortes1995support}
Corinna Cortes and Vladimir Vapnik.
\newblock Support-vector networks.
\newblock \emph{Machine learning}, 20\penalty0 (3):\penalty0 273--297, 1995.

\bibitem[Fisher(1936)]{fisher1936use}
Ronald~A Fisher.
\newblock The use of multiple measurements in taxonomic problems.
\newblock \emph{Annals of eugenics}, 7\penalty0 (2):\penalty0 179--188, 1936.

\bibitem[Forina et~al.(1991)Forina, Leardi, Armanino, and Lanteri]{Forina:1991}
M~Forina, R.~Leardi, C.~Armanino, and S.~Lanteri.
\newblock Wine data set - {UCI} machine learning repository, 1991.
\newblock URL \url{http://archive.ics.uci.edu/ml/datasets/Wine/}.

\bibitem[Hofmann(1994)]{Hofmann:1994}
H.~Hofmann.
\newblock Statlog (german credit data) data set - {UCI} machine learning
  repository, 1994.
\newblock URL
  \url{https://archive.ics.uci.edu/ml/datasets/statlog+(german+credit+data)}.

\bibitem[Huang et~al.(2007)Huang, Ramesh, Berg, and Learned-Miller]{LFWTech}
Gary~B. Huang, Manu Ramesh, Tamara Berg, and Erik Learned-Miller.
\newblock Labeled faces in the wild: A database for studying face recognition
  in unconstrained environments.
\newblock Technical Report 07-49, University of Massachusetts, Amherst, October
  2007.

\bibitem[Jacot et~al.(2018)Jacot, Gabriel, and Hongler]{jacot2018neural}
Arthur Jacot, Franck Gabriel, and Cl{\'e}ment Hongler.
\newblock Neural tangent kernel: Convergence and generalization in neural
  networks.
\newblock In \emph{Advances in neural information processing systems}, pages
  8571--8580, 2018.

\bibitem[Kaynak(1995)]{kaynak1995methods}
C~Kaynak.
\newblock Methods of combining multiple classifiers and their applications to
  handwritten digit recognition.
\newblock \emph{Master’s thesis, Bogazici University}, 1995.

\bibitem[Learned-Miller(2014)]{LFWTechUpdate}
Gary B. Huang~Erik Learned-Miller.
\newblock Labeled faces in the wild: Updates and new reporting procedures.
\newblock Technical Report UM-CS-2014-003, University of Massachusetts,
  Amherst, May 2014.

\bibitem[Lim(1999)]{Lim:1999}
T.-S. Lim.
\newblock Haberman's survival data set - {UCI} machine learning repository,
  1999.
\newblock URL
  \url{https://archive.ics.uci.edu/ml/datasets/Haberman's+Survival}.

\bibitem[Lin and Lin(2003)]{lin2003study}
Hsuan-Tien Lin and Chih-Jen Lin.
\newblock A study on sigmoid kernels for svm and the training of non-psd
  kernels by smo-type methods.
\newblock \emph{submitted to Neural Computation}, 3\penalty0 (1-32):\penalty0
  16, 2003.

\bibitem[Little et~al.(2007)Little, McSharry, Roberts, Costello, and
  Moroz]{little2007exploiting}
Max~A Little, Patrick~E McSharry, Stephen~J Roberts, Declan~AE Costello, and
  Irene~M Moroz.
\newblock Exploiting nonlinear recurrence and fractal scaling properties for
  voice disorder detection.
\newblock \emph{Biomedical engineering online}, 6\penalty0 (1):\penalty0 23,
  2007.

\bibitem[Parisi and RaviChandran(2020)]{parisi2020evolutionary}
Luca Parisi and Narrendar RaviChandran.
\newblock Evolutionary feature transformation to improve prognostic prediction
  of hepatitis.
\newblock \emph{Knowledge-Based Systems}, 200:\penalty0 106012, 2020.

\bibitem[Parisi et~al.(2018{\natexlab{a}})Parisi, RaviChandran, and
  Manaog]{parisi2018decision}
Luca Parisi, Narrendar RaviChandran, and Marianne~Lyne Manaog.
\newblock Decision support system to improve postoperative discharge: A novel
  multi-class classification approach.
\newblock \emph{Knowledge-Based Systems}, 152:\penalty0 1--10,
  2018{\natexlab{a}}.

\bibitem[Parisi et~al.(2018{\natexlab{b}})Parisi, RaviChandran, and
  Manaog]{parisi2018feature}
Luca Parisi, Narrendar RaviChandran, and Marianne~Lyne Manaog.
\newblock Feature-driven machine learning to improve early diagnosis of
  parkinson's disease.
\newblock \emph{Expert Systems with Applications}, 110:\penalty0 182--190,
  2018{\natexlab{b}}.

\bibitem[Parisi et~al.(2020)Parisi, RaviChandran, and Manaog]{parisi2020novel}
Luca Parisi, Narrendar RaviChandran, and Marianne~Lyne Manaog.
\newblock A novel hybrid algorithm for aiding prediction of prognosis in
  patients with hepatitis.
\newblock \emph{Neural Computing and Applications}, 32\penalty0 (8):\penalty0
  3839--3852, 2020.

\bibitem[Patr{\'\i}cio et~al.(2018)Patr{\'\i}cio, Pereira, Cris{\'o}stomo,
  Matafome, Gomes, Sei{\c{c}}a, and Caramelo]{patricio2018using}
Miguel Patr{\'\i}cio, Jos{\'e} Pereira, Joana Cris{\'o}stomo, Paulo Matafome,
  Manuel Gomes, Raquel Sei{\c{c}}a, and Francisco Caramelo.
\newblock Using resistin, glucose, age and bmi to predict the presence of
  breast cancer.
\newblock \emph{BMC cancer}, 18\penalty0 (1):\penalty0 29, 2018.

\bibitem[Pedregosa et~al.(2011)Pedregosa, Varoquaux, Gramfort, Michel, Thirion,
  Grisel, Blondel, Prettenhofer, Weiss, Dubourg, Vanderplas, Passos,
  Cournapeau, Brucher, Perrot, and Duchesnay]{scikit-learn}
F.~Pedregosa, G.~Varoquaux, A.~Gramfort, V.~Michel, B.~Thirion, O.~Grisel,
  M.~Blondel, P.~Prettenhofer, R.~Weiss, V.~Dubourg, J.~Vanderplas, A.~Passos,
  D.~Cournapeau, M.~Brucher, M.~Perrot, and E.~Duchesnay.
\newblock Scikit-learn: Machine learning in {P}ython.
\newblock \emph{Journal of Machine Learning Research}, 12:\penalty0 2825--2830,
  2011.

\bibitem[Rohra et~al.(2017)Rohra, Perumal, Narayanan, Thakur, and
  Bhatt]{rohra2017user}
Jayant~G Rohra, Boominathan Perumal, Swathi~Jamjala Narayanan, Priya Thakur,
  and Rajen~B Bhatt.
\newblock User localization in an indoor environment using fuzzy hybrid of
  particle swarm optimization \& gravitational search algorithm with neural
  networks.
\newblock In \emph{Proceedings of Sixth International Conference on Soft
  Computing for Problem Solving}, pages 286--295. Springer, 2017.

\bibitem[Roweis(2017)]{Roweis:2017}
S.~Roweis.
\newblock Olivetti faces data set, 2017.
\newblock URL \url{https://cs.nyu.edu/~roweis/data.html}.

\bibitem[Rumelhart et~al.(1986)Rumelhart, Hinton, and
  Williams]{rumelhart1986learning}
David~E Rumelhart, Geoffrey~E Hinton, and Ronald~J Williams.
\newblock Learning representations by back-propagating errors.
\newblock \emph{nature}, 323\penalty0 (6088):\penalty0 533--536, 1986.

\bibitem[Vert and Vert(2006)]{vert2006consistency}
R{\'e}gis Vert and Jean-Philippe Vert.
\newblock Consistency and convergence rates of one-class svms and related
  algorithms.
\newblock \emph{Journal of Machine Learning Research}, 7\penalty0
  (May):\penalty0 817--854, 2006.

\bibitem[Wolberg et~al.(1995)Wolberg, Street, and Mangasarian]{Wolberg:1995}
William~H. Wolberg, W.~Nick Street, and Olvi~L. Mangasarian.
\newblock Breast cancer wisconsin (diagnostic) data set - {UCI} machine
  learning repository, 1995.
\newblock URL
  \url{https://archive.ics.uci.edu/ml/datasets/Breast+Cancer+Wisconsin+(Diagnostic)}.

\end{thebibliography}

\end{document}